\documentclass[letterpaper, 10 pt, conference]{ieeeconf}  % Comment this line out if you need a4paper

\IEEEoverridecommandlockouts                              % This command is only needed if 
\usepackage{graphics} % for pdf, bitmapped graphics files
\usepackage{epsfig} % for postscript graphics files
\usepackage{amsmath} % assumes amsmath package installed
\usepackage{amssymb,bm}  % assumes amsmath package installed
\usepackage{color}
\usepackage[style=ieee]{biblatex}

\DeclareSourcemap{
  \maps{
    \map{
      \pertype{article}
      \step[fieldset=url, null]
      \step[fieldset=doi, null]
      \step[fieldset=issn, null]
      \step[fieldset=isbn, null]
      \step[fieldset=note, null]
      \step[fieldset=editor, null]
      \step[fieldset=urldate, null]
      \step[fieldset=file, null]
    }
  }
}
\DeclareSourcemap{
  \maps{
    \map{
      \pertype{inproceedings}
      \step[fieldset=url, null]
      \step[fieldset=doi, null]
      \step[fieldset=issn, null]
      \step[fieldset=isbn, null]
      \step[fieldset=note, null]
      \step[fieldset=editor, null]
      \step[fieldset=urldate, null]
      \step[fieldset=file, null]
    }
  }
}
\DeclareSourcemap{
  \maps{
    \map{
      \pertype{incollection}
      \step[fieldset=url, null]
      \step[fieldset=doi, null]
      \step[fieldset=issn, null]
      \step[fieldset=isbn, null]
      \step[fieldset=note, null]
      \step[fieldset=editor, null]
      \step[fieldset=urldate, null]
      \step[fieldset=file, null]
    }
  }
}

\title{\LARGE \bf
Capture Point Control in Thruster-Assisted Bipedal Locomotion 
}

\author{Shreyansh Pitroda \textsuperscript{1\textdagger}, Aditya Bondada\textsuperscript{1\textdagger}, Kaushik Venkatesh$^{1}$, Adarsh Salagame$^{1}$,  Chenghao Wang$^{1}$,\\ Taoran Liu$^{1}$, Bibek Gupta$^{1}$, Eric Sihite$^{2}$, Reza Nemovi$^{2}$, Alireza Ramezani$^{1 *}$, and Morteza Gharib$^{2}$% <-this % stops a space
\thanks{$^{1}$ The authors are with the Department of Electrical Engineering, Northeastern University, USA.}%
\thanks{$^{2}$ The authors are with the Department of Aerospace Engineering, California Institute of Technology, USA.}%
\thanks{ \textsuperscript{\textdagger} These authors have equal contribution to this work}
\thanks{$^{*}$ The corresponding author, email: a.ramezani@northeastern.edu}
}

\begin{document}

\maketitle
\thispagestyle{empty}
\pagestyle{empty}

%%%%%%%%%%%%%%%%%%%%%%%%%%%%%%%%%%%%%%%%%%%%%%%%%%%%%%%%%%%%%%%%%%%%%%%%%%%%%%%%
\begin{abstract}
%
%\color{blue}
Despite major advancements in control design that are robust to unplanned disturbances, bipedal robots are still susceptible to falling over and struggle to negotiate rough terrains. By utilizing thrusters in our bipedal robot, we can perform additional posture manipulation and expand the modes of locomotion to enhance the robot's stability and ability to negotiate rough and difficult-to-navigate terrains. In this paper, we present our efforts in designing a controller based on capture point control for our thruster-assisted walking model named Harpy and explore its control design possibilities. While capture point control based on centroidal models for bipedal systems has been extensively studied, the incorporation of external forces that can influence the dynamics of linear inverted pendulum models, often used in capture point-based works, has not been explored before. The inclusion of these external forces can lead to interesting interpretations of locomotion, such as virtual buoyancy studied in aquatic-legged locomotion. This paper outlines the dynamical model of our robot, the capture point method we use to assist the upper body stabilization, and the simulation work done to show the controller's feasibility.
\end{abstract}

%%%%%%%%%%%%%%%%%%%%%%%%%%%%%%%%%%%%%%%%%%%%%%%%%%%%%%%%%%%%%%%%%%%%%%%%%%%%%%%%
\section{Introduction and Motivation}

Raibert's robots \cite{murphy_littledog_2011} and Boston Dynamics' robots \cite{noauthor_robots_nodate} represent some of the most successful examples of legged robots, demonstrating robust hopping or trotting capabilities even in the face of significant unplanned disturbances. Alongside these achievements, numerous underactuated and fully actuated bipedal robots have been introduced \cite{park_finite-state_2013,buss_preliminary_2014,ramezani_performance_2014,grizzle_progress_nodate}. Agility Robotics' Cassie \cite{apgar_fast_2018} and Hubo \cite{wang_drc-hubo_2014} exhibit capabilities ranging from walking and running to dancing and navigating stairs, while Atlas recover from pushes \cite{koolen_design_2016}.

Despite these advancements, all these systems remain susceptible to falling over and struggle to negotiate extremely rough terrains. Even humans, renowned for their natural, dynamic, and robust gaits, cannot consistently recover from severe terrain perturbations, external pushes, or slips on icy surfaces. Our objective is to enhance the robustness of these systems by implementing a distributed array of thrusters and employing nonlinear control techniques.

The application of thrusters (thrust vectoring) and posture manipulation has recently undergone testing in notable examples such as the Multi-modal mobility morphobot (M4) \cite{sihite_multi-modal_2023,sihite_efficient_2022,mandralis_minimum_2023} and LEONARDO \cite{kim_bipedal_2021,liang_rough-terrain_2021,sihite_optimization-free_2021,sihite_efficient_2022}. M4 endeavors to enhance its locomotion versatility by integrating posture manipulation and thrust-vectoring to increase the variety of locomotion modes. Conversely, LEONARDO is a legged robot equipped with a multitude of propellers, enabling both walking and flying capabilities. However, neither of these examples adequately demonstrates dynamic legged locomotion and aerial mobility, which presents a formidable challenge due to conflicting requirements inherent in these modes of operation. The integration of these modes into a single platform remains a significant hardware design obstacle.

\begin{figure}[t]
    \vspace{0.05in}
    \centering
    \includegraphics[width=0.7\linewidth]{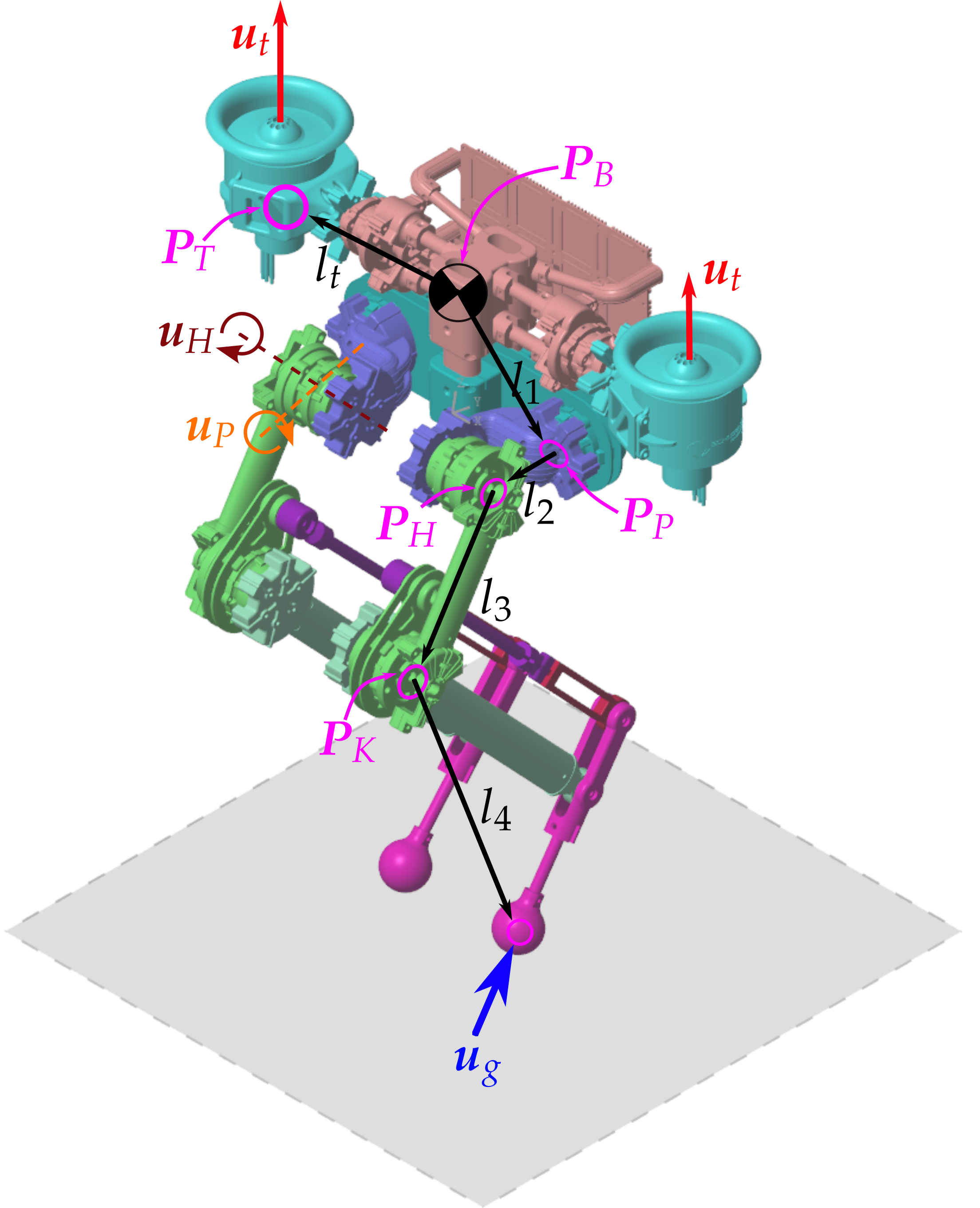}
    \caption{Illustrates the CAD model of Harpy platform, a bipedal robot with two electric ducted fans attached to its torso.}
    \label{fig:cover-image}
    \vspace{-0.05in}
\end{figure}

Posture manipulation and thrust-vectoring are commonly used in birds, notable examples are chuckar birds which are capable of showcasing wing-assisted incline running maneuver known as WAIR problem \cite{dial_wing-assisted_2003-1,tobalske_aerodynamics_2007}. In WAIR maneuver, Chukar birds apply their flapping wings and resulting aerodynamic forces to enhance contact forces to walk up steep slopes.

In this paper, we present our efforts in designing a controller based on capture point control for our thruster-assisted walking model named \textit{Harpy} (depicted in Fig.~\ref{fig:cover-image}). With a total of eight actuators and a pair of electric ducted fans fixed to its torso, this biped aims to combine the mobility advantages of aerial and legged systems, capable of achieving dynamic terrestrial locomotion and flight. The hardware design and assembly of Harpy have been completed, and the primary objective of this work is to explore control design possibilities for Harpy \cite{dangol_feedback_2020-1,dangol_control_2021}.

While capture point control based on centroidal models for bipedal systems has been extensively studied \cite{koolen_design_2016}, the incorporation of external forces that can influence the dynamics of linear inverted pendulum models, often used in capture point-based works, has not been explored before. The inclusion of these external forces can lead to interesting interpretations of locomotion, such as virtual buoyancy studied in aquatic legged locomotion.

In this work, we consider these external forces—in the form of thruster forces—which can be utilized to modulate the solutions within the elliptical energy that define capture point solutions. These adjustments reduce the effort required by the system (e.g., step length for recovery) to respond to external perturbations, thereby enhancing overall robustness in scenarios prone to tipping over. The primary contribution of this work lies in taking meaningful steps towards the unexplored domain of thruster-assisted dynamic terrestrial locomotion. 

This work is structured as follows: we present the derivations of Harpy reduced order model, followed by the capture point control, simulation results, and concluding remarks.

\section{Harpy Reduced-Order Model (HROM) Derivations}

This section outlines the dynamics formulation of the robot which is used in the numerical simulation in Section 3, in addition to the reduced order models which are used in the controller design. Figure \ref{fig:cover-image} shows the kinematic configuration of Harpy which listed the center of mass (CoM) positions of the dynamic components, joint actuation torques, and thruster torques. The system model has a combined total of 12 degrees-of-freedoms (DoFs): 6 for the body and 3 on each leg. Due to the symmetry, the left and right side of the robot follow a similar derivations so only the general derivations are provided in this section.

\subsection{Energy-based Lagrange Formalism}

The Harpy equations of motion are derived using Euler-Lagrangian dynamics formulation. In order to simplify the system, each linkage is assumed to be massless, with the mass concentrated at the body and the joint motors. Consequently, the lower leg kinematic chain is considered massless, significantly simplifying the system. The three leg joints are labeled as the hip frontal (pelvis $P$), hip sagittal (hip $H$), and knee sagittal (knee $K$), as illustrated in Fig. \ref{fig:cover-image}. The thrusters are also considered massless and capable of providing forces in any direction to simplify the problem.

Let $\gamma_h$ be the frontal hip angle, while $\phi_h$ and $\phi_k$ represent the sagittal hip and knee angles, respectively. The superscripts $\{B,P,H,K\}$ represent the frame of reference about the body, pelvis, hip, and knee, while the inertial frame is represented without the superscript. Let $R_B$ be the rotation matrix from the body frame to the inertial frame (i.e., $\bm x = R_B\, \bm x^B$). The pelvis motor mass is added to the body mass. Then, the positions of the hip and knee centers of mass (CoM) are defined using kinematic equations:
\begin{equation}
\begin{gathered}
    \bm{p}_P = \bm{p}_{B} + R_{B}\, \bm{l}_{1}^{B}, \\
    \bm{p}_H = \bm{p}_{P} + R_{B}\,R_x(\gamma_h)\, \bm{l}_{2}^{P} \\
    \bm{p}_K = \bm{p}_{H} + R_{B}\,R_x(\gamma_h)\,R_y(\phi_h) \bm{l}_{3}^{H},
\end{gathered}
\label{eq:pos_com}
\end{equation}
where $R_x$ and $R_y$ are the rotation matrices about the $x$ and $y$ axes, respectively, and $\bm l$ is the length vector representing the configuration of Harpy, which remains constant in its respective local frame of reference. The positions of the foot and thrusters are defined as:
\begin{equation}
\begin{gathered}
    \bm{p}_F = \bm{p}_{K} + R_{B}\,R_x(\gamma_h)\,R_y(\phi_h)\,R_y(\phi_k)\, \bm{l}_{4}^{K} \\
    \bm{p}_T = \bm{p}_{B} + R_{B}\, \bm{l}_{t}^{B}
\end{gathered}
\label{eq:pos_other}
\end{equation}
where the length vector from the knee to the foot is $\bm l_4^K = [-l_{4a}\cos{\phi_k}, 0, -( l_{4b} + l_{4a}\sin{\phi_k})]^\top$, which represents the kinematic solution to the parallel linkage mechanism of the lower leg. Let $\bm \omega_B$ be the angular velocity of the body. Then, the angular velocities of the hip and knee are defined as: $\bm \omega_H^B = [\dot{\gamma}_h,0,0]^\top + \bm \omega_B^B$ and $\bm \omega_K^H = [0,\dot{\phi}_h,0]^\top + \bm \omega_H^H$. Consequently, the total energy of Harpy for the Lagrangian dynamics formulation is defined as follows:
\begin{equation}
\begin{aligned}
    K &= \tfrac{1}{2} \textstyle \sum_{i \in \mathcal{F}} \left( 
        m_i\,\bm p_i^\top\, \bm p_i + 
        \bm \omega_i^{i \top} \, \hat I_i \, \bm \omega_i^i \right) \\
    V &= - \textstyle \sum_{i \in \mathcal{F}} \left( 
        m_i\,\bm p_i^\top\, [0,0,-g]^\top \right),
\end{aligned}
\label{eq:energy}
\end{equation}
where $\mathcal{F} = \{B,H_L,K_L,H_R,K_R\}$ represents the relevant frames of reference and mass components (body, left hip, left knee, right hip, right knee), and the subscripts $L$ and $R$ denote the left and right sides of the robot, respectively. Furthermore, $\hat I_i$ denotes the inertia about its local frame, and $g$ is the gravitational constant. This constitutes the Lagrangian of the system, given by $L = K - V$, which is utilized to derive the Euler-Lagrange equations of motion. The dynamics of the body's angular velocity are derived using the modified Lagrangian for rotation in $SO(3)$ to avoid using Euler angles and the potential gimbal lock associated with them. This yields the following equations of motion following Hamilton's principle of least action:
\begin{equation}
\begin{gathered}
    \tfrac{d}{dt}\left( \tfrac{\partial L}{\partial \bm \omega_B^B}  \right) + 
    \bm \omega_B^B \times \tfrac{\partial L}{\partial \bm \omega_B^B} + 
    \textstyle \sum_{j=1}^{3} \bm r_{Bj} \times \tfrac{\partial L}{\partial \bm r_{Bj}} = \bm u_1, \\
    \tfrac{d}{dt}\left( \tfrac{\partial L}{\partial \dot {\bm q}}  \right) - 
    \tfrac{\partial L}{\partial \bm q} = \bm u_2, \\ 
    \tfrac{d}{dt} R_B = R_B\, [\bm \omega_B^B]_\times,
\end{gathered}
\label{eq:eom_eulerlagrange}
\end{equation}
where $[\, \cdot \, ]_\times$ denotes the skew symmetric matrix, $R_B^\top = [\bm r_{B1}, \bm r_{B2}, \bm r_{B3}]$, $\bm q = [\bm p_B^\top, \gamma_{h_L}, \gamma_{h_R}, \phi_{h_L}, \phi_{h_R}]^\top$ represents the dynamical system states other than $(R_B,\bm \omega^B_B)$, and $\bm u$ denotes the generalized forces. The knee sagittal angle $\phi_k$, which is not associated with any mass, is updated using the knee joint acceleration input $\bm u_k = [\ddot{\phi}_{k_L}, \ddot{\phi}_{k_R}]^\top$. Then, the system acceleration can be derived as follows:
\begin{equation}
\begin{gathered}
    M \bm a + \bm h = B_j\, \bm u_j + B_t\, \bm u_t + B_g\, \bm u_g
\end{gathered}
\label{eq:eom_accel}
\end{equation}
where $\bm a = [ \dot{\bm \omega}_B^{B\top}, \ddot{\bm q}^\top, \ddot{\phi}_{k_L}, \ddot{\phi}_{k_R}]^\top$, $\bm u_t$ denotes the thruster force, $\bm u_j = [u_{P_L}, u_{P_R}, u_{H_L}, u_{H_R}, \bm u_k^\top]^\top$ represents the joint actuation, and $\bm u_g$ stands for the ground reaction forces (GRFs). The variables $M$, $\bm h$, $B_t$, and $B_g$ are functions of the full system states:
\begin{equation}
    \bm x = [\bm r_{B}^\top, \bm q^\top, \phi_{K_L}, \phi_{K_R}, \bm \omega_B^{B \top}, \dot{\bm q}^\top, \dot{\phi}_{K_L}, \dot{\phi}_{K_R}]^\top,
\label{eq:states}
\end{equation}
where the vector $\bm r_B$ contains the elements of $R_B$. Introducing $B_j = [0_{6 \times 6}, I_{6 \times 6}]$ allows $\bm u_j$ to actuate the joint angles directly. Let $\bm v = [\bm \omega_B^{B\top}, \dot{\bm q}^\top]^\top$ denote the velocity of the generalized coordinates. Then, $B_t$ and $B_g$ can be defined using the virtual displacement from the velocity as follows:
\begin{equation}
\begin{aligned}
    B_t = \begin{bmatrix}
        \begin{pmatrix}
        \partial \dot{\bm p}_{T_L} / \partial \bm v \\
        \partial \dot{\bm p}_{T_R} / \partial \bm v
        \end{pmatrix}^\top
        \\
        0_{2 \times 6}
    \end{bmatrix}, \quad
    B_g = \begin{bmatrix}
        \begin{pmatrix}
        \partial \dot{\bm p}_{F_L} / \partial \bm v \\
        \partial \dot{\bm p}_{F_R} / \partial \bm v
        \end{pmatrix}^\top
        \\
        0_{2 \times 6}
    \end{bmatrix}.
\end{aligned}
\label{eq:generalized_forces}
\end{equation}
The vector $\bm u_t = [\bm u_{t_L}^\top, \bm u_{t_R}^\top]^\top$ is composed of the left and right thruster forces $\bm u_{t_L}$ and $\bm u_{t_R}$, respectively. The GRF is modeled using the unilateral compliant ground model with undamped rebound, while friction is modeled using the Stribeck friction model, defined as follows:
\begin{equation}
\begin{aligned}
    u_{g,z} =& -k_{g,p}\, p_{F,z} - k_{g,d}\, \dot p_{F,z} \\
    u_{g,x} =& -\left(\mu_c + (\mu_s - \mu_c)\, \mathrm{exp}\left(-\tfrac{|\dot p_{F,x}|^2}{v_s^2}\right) \right) f_z\, \mathrm{sgn}(\dot p_{F,x}) \\ 
    & - \mu_v\,\dot p_{F,x},
\end{aligned}
\label{eq:ground_model}
\end{equation}
where $p_{F,x}$ and $p_{F,z}$ represent the $x$ and $z$ components of the inertial foot position, $k_{g,p}$ and $k_{g,d}$ denote the spring and damping model for the ground, $\mu_c$, $\mu_s$, and $\mu_v$ are the Coulomb, static, and viscous friction coefficients, respectively, and $v_s$ is the Stribeck velocity. $k_{g,d}$ is set to $0$ if $\dot p_{F,z} > 0$ for the undamped rebound model, and friction in the $y$ direction follows a similar derivation to $u_{g,x}$. Then, the ground force model $\bm u_g$ is defined as follows:
\begin{equation}
\begin{aligned}
    \bm u_g = [\bm u_{g_L}^\top\, H(-p_{F_L,z}),\, \bm u_{g_R}^\top\, H(-p_{F_R,z})]^\top,
\end{aligned}
\label{eq:ground_forces}
\end{equation}
where $H(x)$ denotes the Heaviside function, while $\bm u_{g_L}$ and $\bm u_{g_R}$ represent the left and right ground forces, which are formed using their respective components $u_{g,x}$, $u_{g,y}$, and $u_{g,z}$.

The full-dynamics model can be derived using equations \eqref{eq:eom_eulerlagrange} to \eqref{eq:ground_forces} to form $\dot{\bm x} = \bm f(\bm x, \bm u_j, \bm u_t, \bm u_g)$. Finally, using the full-dynamics derived above, we proceed to ROM derivations. As shown in Fig. \ref{fig:hrom}, the model is described using the inverted pendulum model, where the length of $r$ can be adjusted through the change in leg conformation, i.e., variable-length inverted pendulum model (VLIP).

In the VLIP model, the center of pressure (CoP), denoted as $\bm c$, is defined as the weighted average position of the feet, given by $\bm c = \lambda_L\, \bm p_{F_L} + \lambda_R\, \bm p_{F_R}$, where $\lambda_i = u_{g_i,z} / (u_{g_L,z} + u_{g_R,z})$ for $i \in \{L,R\}$. In the Harpy full-dynamics model, which uses a point foot, $\bm c$ equals the stance foot position during the SS phase. The VLIP model without thrusters is underactuated, but the addition of thrusters makes the system fully actuated and enables trajectory tracking. Hence, the VLIP model is derived as follows:
\begin{equation}
\begin{gathered}
    m \ddot{\bm p}_B = m \bm g + \bm u_{t,c} + J_s^\top \bm \lambda\\ 
\end{gathered}
\label{eq:model_vlip}
\end{equation}
where $m$ represents the mass of the VLIP model, which in this case is the total mass of the system, and $\bm u_{t,c}$ denotes the thruster forces about the CoM. The constraint force $J_s^\top \bm \lambda$ is established to maintain the leg length $r$ equal to the leg conformation, utilizing the following constraint equation:
\begin{equation}
\begin{gathered}
    J_s\, (\ddot{\bm p}_B - \ddot {\bm c}) = u_r, \\  
    J_s = (\bm p_B - \bm c)^\top,
\end{gathered}
\label{eq:model_vlip_constraint}
\end{equation}
which is designed to maintain the leg length's second derivative equal to $u_r$. This constraint force also constitutes the GRF as long as the friction cone constraint is satisfied. Assuming no slip ($\ddot{\bm c} = 0$), the inputs to the system are $u_r$, which controls the body position about the vector $\bm r = \bm p_B - \bm c$ by adjusting the leg length, and the thrusters $\bm u_t$, which control the remaining degrees of freedom.

\section{Capture Point Control}

\begin{figure}[t]
    \centering
    \includegraphics[width=\linewidth]{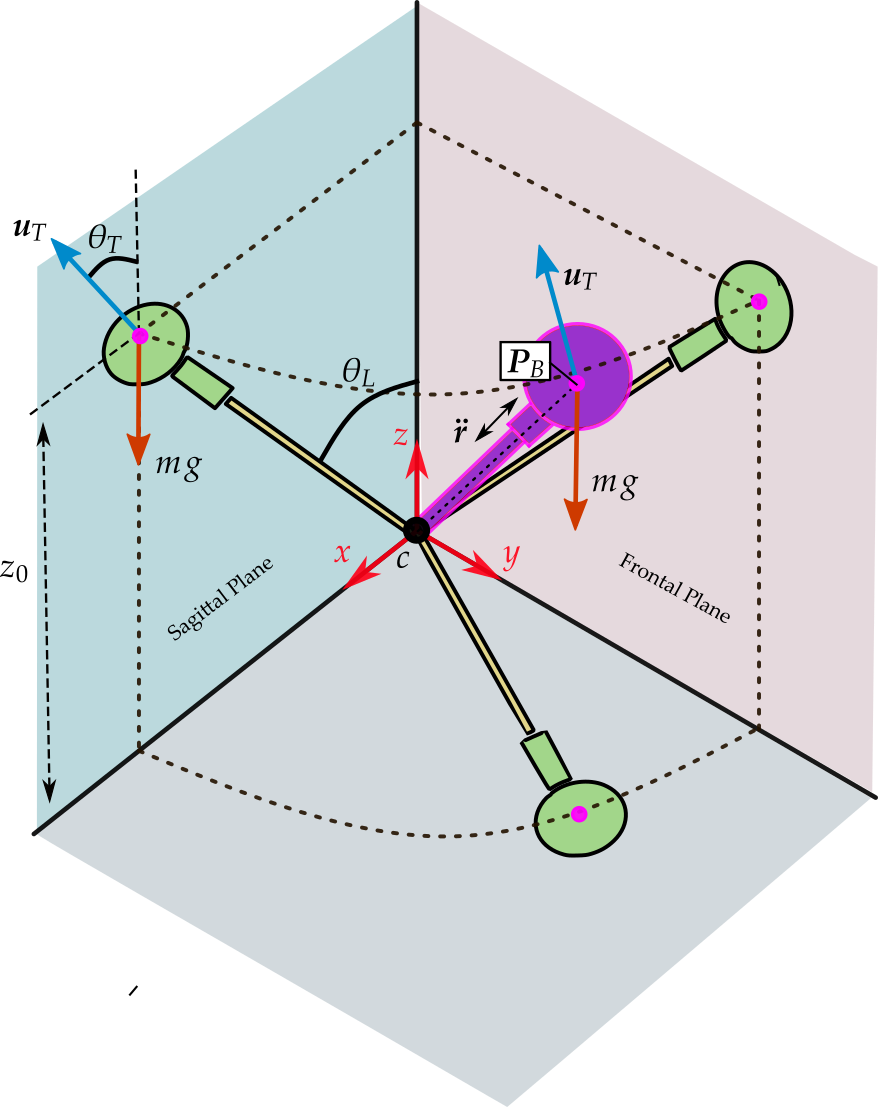}
    \caption{Illustrates Harpy reduced-order model parameters. A variable-length inverted pendulum model with thruster force is projected to the sagittal, frontal and transversal planes of locomotion.}
    \label{fig:hrom}
\end{figure}

\begin{figure}
    \centering
    \includegraphics[width=1\linewidth]{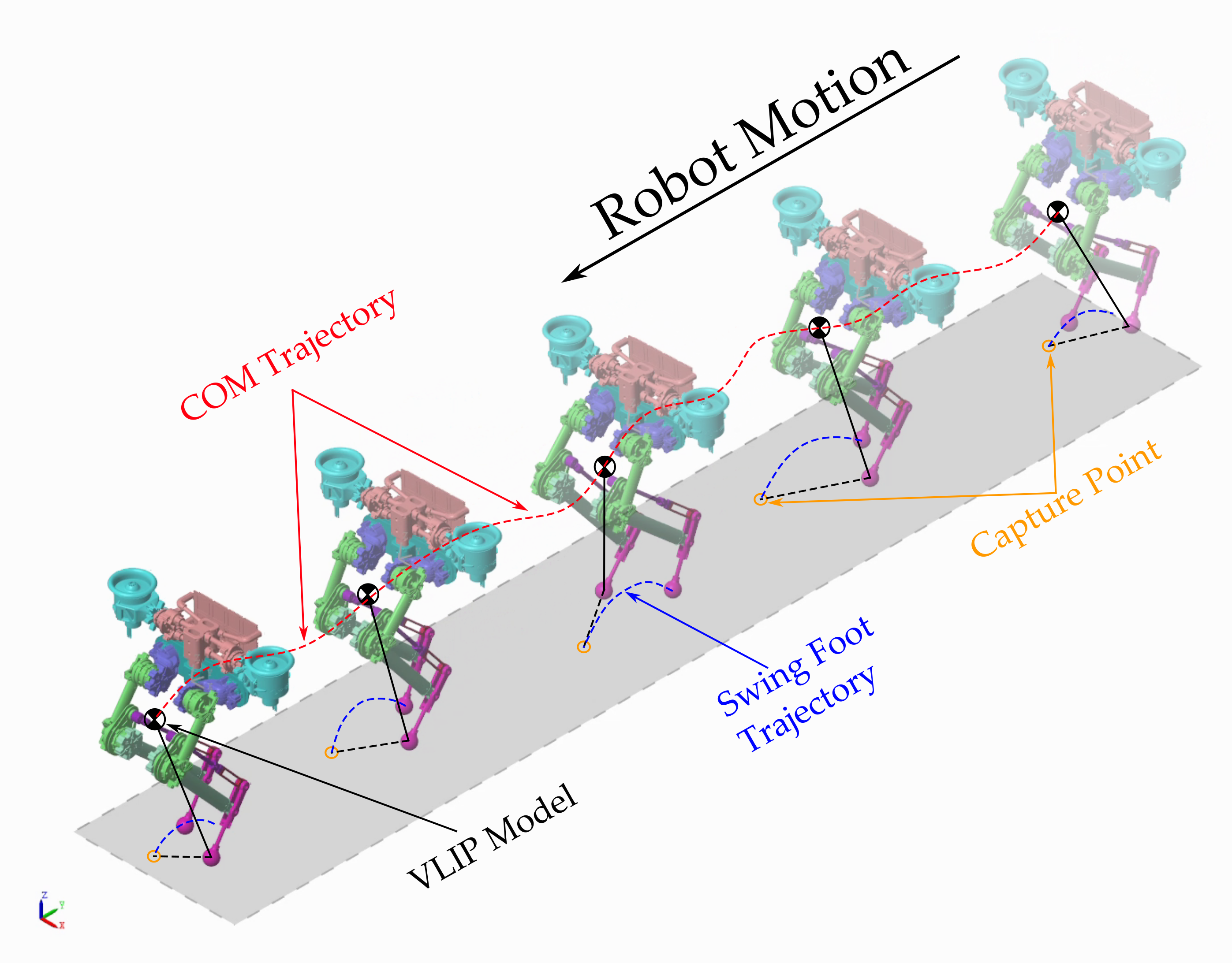}
    \caption{Snap shots of Harpy's thruster-assisted walking on flat ground.}
    \label{fig:harpy-walking-snapshots}
\end{figure}

\begin{figure}
    \centering
    \includegraphics[width=\linewidth]{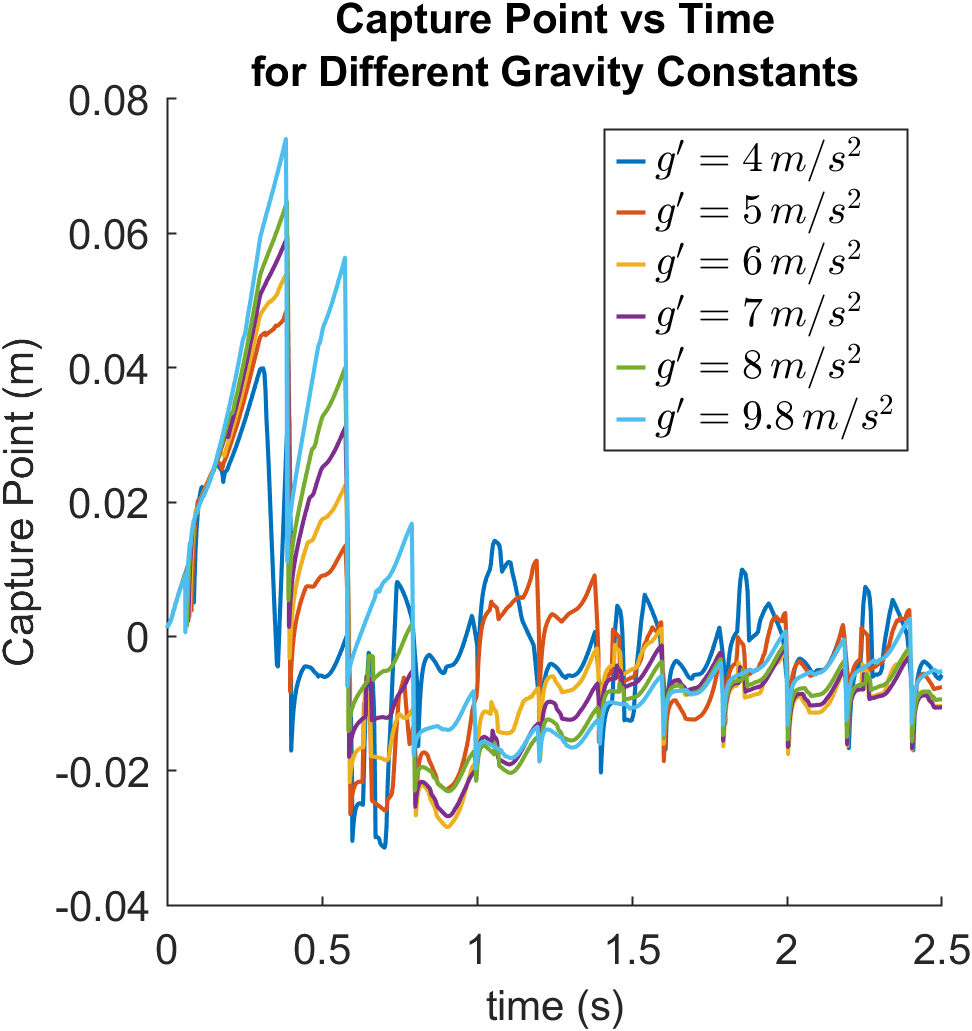}
    \caption{ Capture point obtained for various $g'=g-\bm u_{t,c}/m$.}
    \label{fig:cp}
\end{figure}

\begin{figure*}
    \centering
    \includegraphics[width=0.8\linewidth]{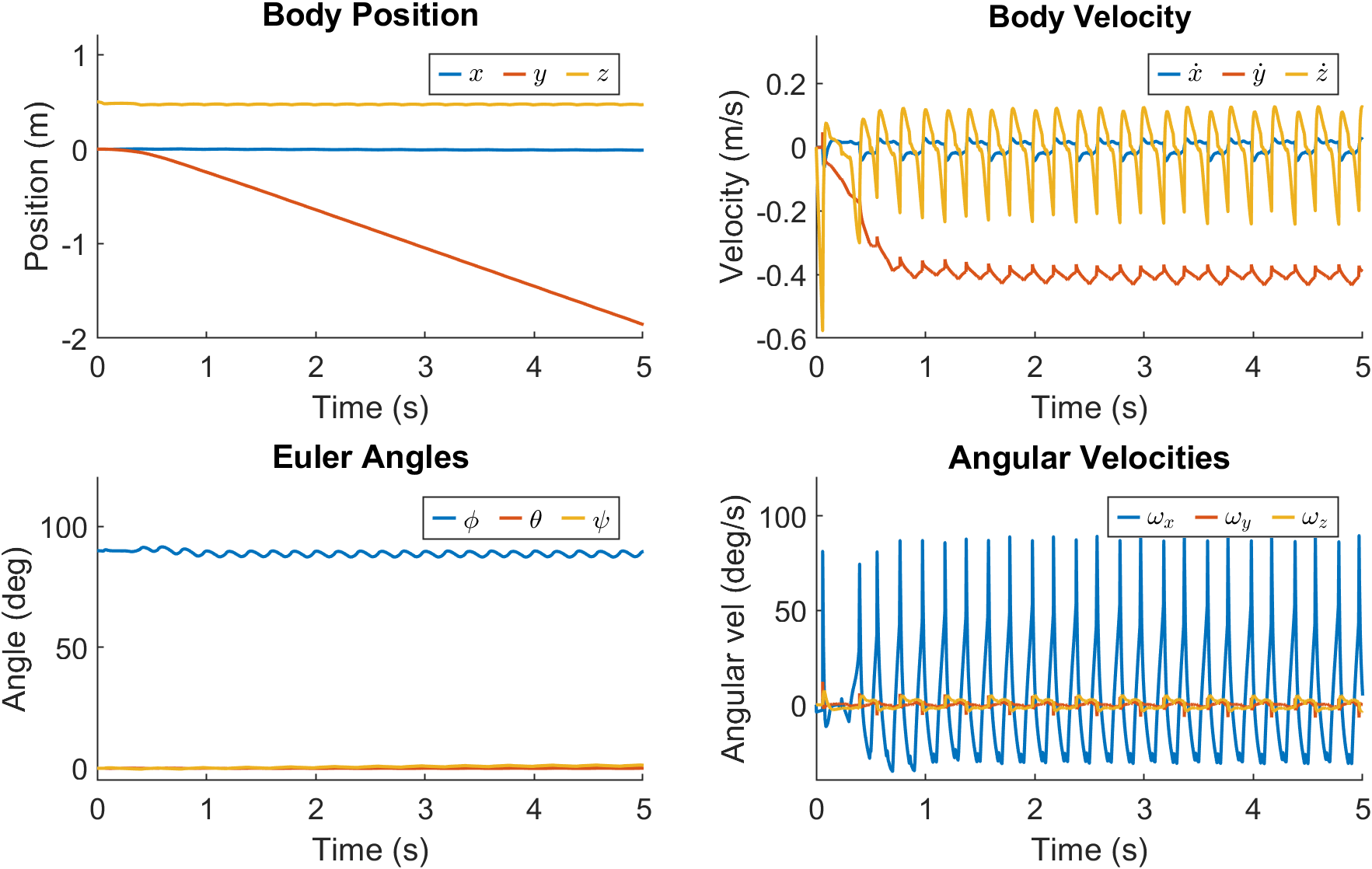}
    \caption{Illustrates Harpy state variable trajectories evolution obtained using the Simscape/Matlab model.}
    \label{fig:states}
\end{figure*}

\begin{figure}
    \centering
    \includegraphics[width=1\linewidth]{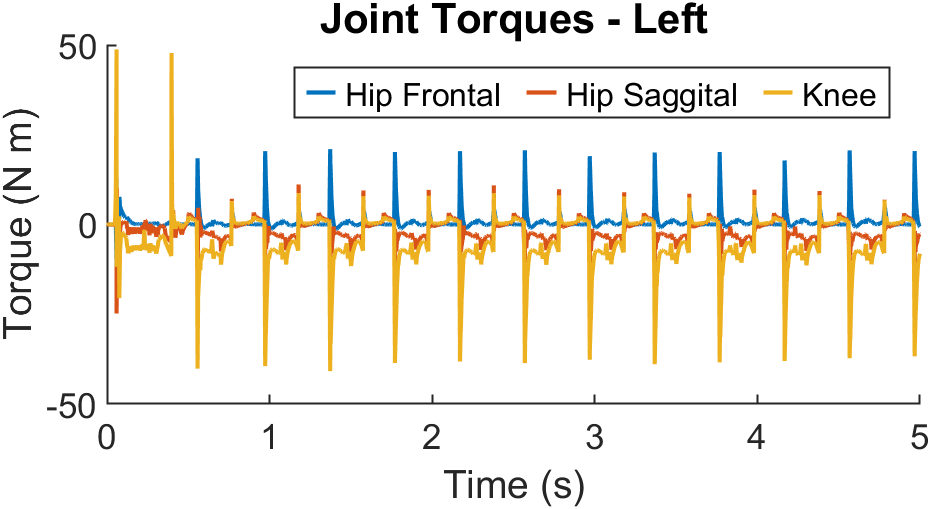}
    \caption{Illustrates Harpy joint torques obtained using the Simscape/Matlab model.}
    \label{fig:torques}
\end{figure}
\begin{figure}
    \centering
    \includegraphics[width=1\linewidth]{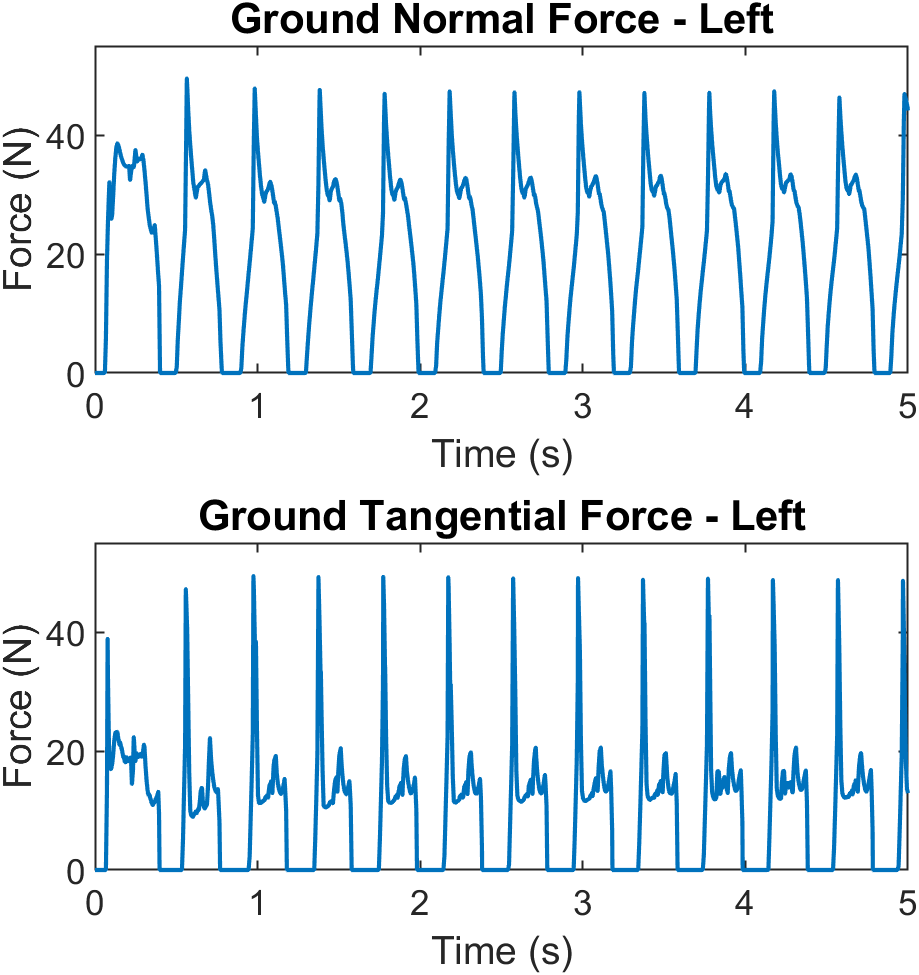}
    \caption{Illustrates Harpy ground reaction forces (GRF) obtained using the Simscape/Matlab model.}
    \label{fig:grf}
\end{figure}

\begin{figure}
    \centering
    \includegraphics[width=1\linewidth]{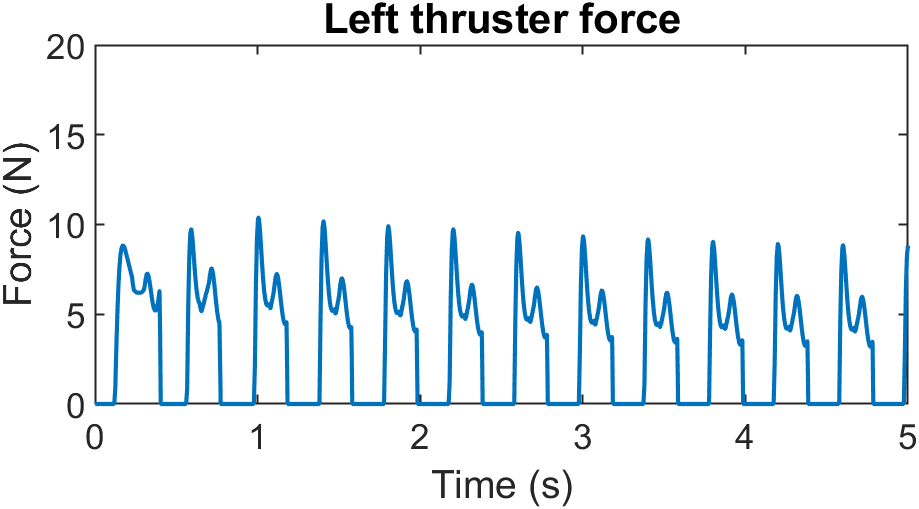}
    \caption{Illustrates thruster commands time-evolution obtained using the Simscape/Matlab model.}
    \label{fig:thruster}
\end{figure}

\begin{figure}
    \centering
    \includegraphics[width=1\linewidth]{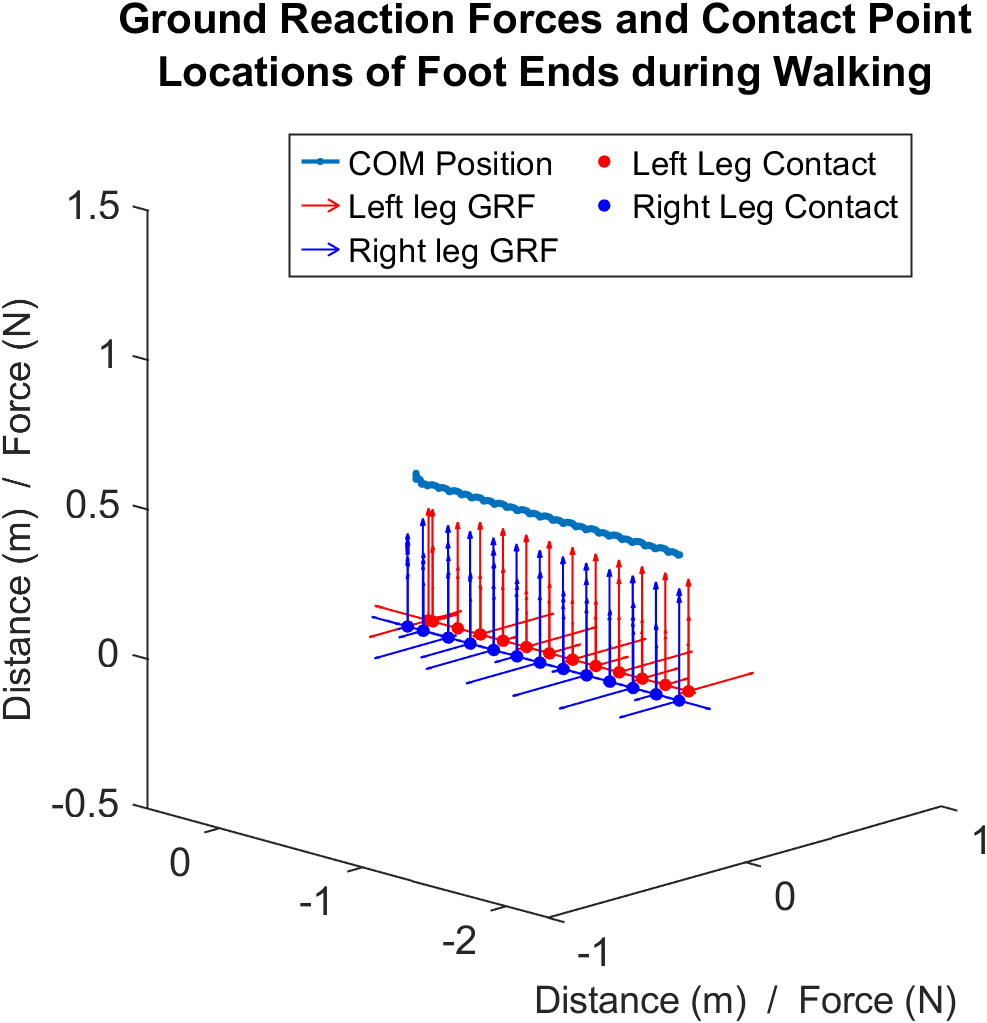}
    \caption{Illustrates contact points, unilateral force vectors, and CoM trajectory in an isometric view.}
    \label{fig:grf-contact-pnt-iso}
\end{figure}

\begin{figure}
    \centering
    \includegraphics[width=1\linewidth]{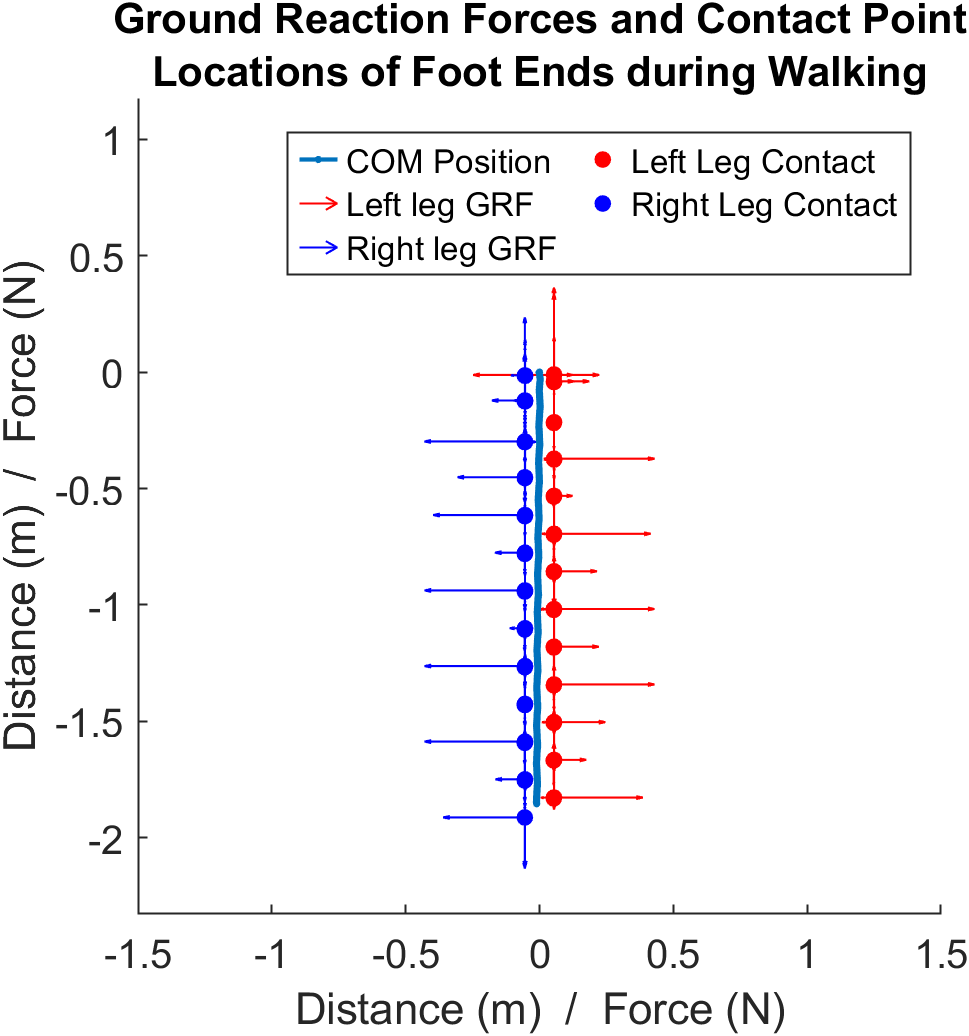}
    \caption{Illustrates contact points, unilateral force vectors, and CoM trajectory in a top view.}
    \label{fig:grf-contact-pnt-top}
\end{figure}

% \begin{figure}[h]
%     \vspace{0.05in}
%     \centering
%     \includegraphics[width=0.8\linewidth]{images/Capture-point-params.png}
%     \caption{ Harpy's VLIP model}
%     \label{fig:}
%     \vspace{-0.05in}
% \end{figure}

To design the controller, as depicted in Fig.~\ref{fig:hrom}, we consider the projection of the VLIP model onto the sagittal and frontal planes of locomotion and design the capture point controller separately. Here, we elucidate the control design for the sagittal plane. We start with a biped system abstracted as a planar inverted pendulum. The equations of motion in the $x$-$z$ plane are given by
\begin{equation}
\begin{aligned}
& m \ddot{p}_{B,x}=|\bm\lambda| \sin \theta_L+\left|\bm u_{t, c}\right| \sin \theta_T \\
& m \ddot{p}_{B,z}=-mg+|\bm \lambda| \cos \theta_L+\left|\bm u_{t, c}\right| \cos \theta_T
\end{aligned}
\label{eq:sagittal-vlip}
\end{equation}
where $\theta_L$ and $\theta_T$ are illustrated in Fig.~\ref{fig:hrom}. The linear pendulum model can be enforced by setting $p_{B,z}=z_0$ and $\ddot p_{B,z}=0$. Therefore, the magnitude of $\bm \lambda$ is determined by
\begin{equation}
    \left|\bm \lambda\right|=\left(mg-\left|\bm u_{t, c}\right| \cos \theta_T\right) \frac{\left|\bm r\right|}{z_0}
\end{equation}
By substituting $\sin \theta_L=\frac{x}{r}$ and $\left|\bm \lambda\right|$ from above into Eq.~\ref{eq:sagittal-vlip}, $\ddot p_{B,x}$ is given by
\begin{equation}
    m\ddot{p}_{B,x}=\frac{x}{z_0}\left(mg-\left|\bm u_{t, c}\right| \cos \theta_T\right)+\left|\bm u_{t, c}\right| \sin \theta_T
\end{equation}
Note that if through torso angle manipulation thruster actions around the CoM $\bm u_{t,c}$ are kept perpendicular to the ground surface, i.e., $\theta_T=0$, then we can express the virtual mass-spring model with a negative stiffness rate $-\left(g-\frac{\left|\bm u_{t,c}\right|}{m}\right)$ as follows:
\begin{equation}
    \ddot{p}_{B,x}=\left(g-\frac{\left|\bm u_{t,c}\right|}{m}\right) \frac{p_{B,x}}{z_0}
\end{equation}
Since the stiffness rate in this model is negative and dictated by the thrusters, we refer to this model as virtual buoyancy. It is possible to observe that the thruster force can reduce the walking frequency, similar to submersed aquatic-legged locomotion. The orbital energy $E$ of the virtual buoyancy model is given by 
\begin{equation}
    E=\frac{1}{2} \dot{p}_{B,x}^2-\frac{1}{2}\left(g-\frac{\left|\bm u_{t, c}\right|}{m}\right) \frac{p_{B,x}^2}{z_0}
\end{equation}
When the CoM moves towards the foot and $E > 0$, there is sufficient energy for the CoM to pass over the foot and maintain its motion. Conversely, if $E < 0$, the CoM halts and changes direction before reaching over the foot. At $E = 0$, the CoM comes to a rest directly above the foot. This equilibrium state, $E = 0$, defines the two eigenvectors of the buoyancy model, expressed as:
\begin{equation}
    \dot{p}_{B,x}= \pm p_{B,x} \sqrt{\frac{g-\frac{\left|\bm u_{t,c}\right|}{m}}{z_0}}
\end{equation}
The equation above depicts a saddle point characterized by one stable and one unstable eigenvector. In the stable eigenvector, $p_{B,x}$ and $\dot p_{B,x}$ exhibit opposite signs, indicating that the CoM is approaching the CoP. Conversely, in the unstable eigenvector, they share the same signs, indicating that the CoM is moving away from the CoP. The orbital energy of the inverted pendulum remains constant until the swing leg is placed and the roles of the feet are exchanged. Assuming this exchange occurs instantaneously without energy loss, we can determine the foot placement based on the capture point, given by
\begin{equation}
    p_{B,x}=\dot p_{B,x} \sqrt{\frac{z_0}{g-\frac{\left|\bm u_{t,c}\right|}{m}}}
\end{equation}

\section{Results and Discussion}
\label{sec:results}

The implementation of the capture point control strategy for Harpy's thruster-assisted walking model yielded promising results in simulations. A high-fidelity model of Harpy was developed in Simscape, as depicted in Fig.~\ref{fig:harpy-walking-snapshots}. The controller quickly attained a stable limit cycle, as shown in Fig.~\ref{fig:states}, indicating the system's ability to maintain stable walking motion using our controller. Figure~\ref{fig:torques} illustrates the initial joint torque for the left knee, which was high as Harpy was dropped from slightly above ground at the beginning of the simulation. Ground reaction forces are shown in Fig.~\ref{fig:grf}. The thruster force, shown in Fig.~\ref{fig:thruster}, was calculated from the controller. This thruster force was then integrated into the capture point controller, contributing to the virtual buoyancy model. The thruster contributions reduced the effort required for fallover prevention as observed in Fig.~\ref{fig:cp}. Additionally, Figs.~\ref{fig:grf-contact-pnt-iso} and \ref{fig:grf-contact-pnt-top} demonstrate Harpy's contact forces, stance foot locations, and CoM trajectory in isometric and top views.

\section{Conclusion}
\label{sec:conclusion}

We presented the design and implementation of a thruster-assisted walking controller for Harpy, a biped robot. The high-fidelity model of Harpy allowed for a detailed analysis of the system's dynamics and control strategies. We employed a control design based on capture point theory, which identifies a region for foot placement such that the overall energy of the CoM dissipates to halt the system, preventing potential fallover situations. Our controller demonstrated the ability to quickly attain a stable limit cycle, indicating its effectiveness.

Future work will focus on experimental validation of our controller on the hardware of Harpy, which has recently been completed. Additionally, further improvements to the controller algorithms and integration of various perception elements, such as vision feedback, will be pursued to enhance Harpy's robustness and adaptability in complex terrain. Overall, our results suggest that thruster-assisted walking has the potential to offer a fresh perspective on legged locomotion and provide rich opportunities for unexplored control design avenues.

\printbibliography

@inproceedings{mandralis_minimum_2023,
	title = {Minimum {Time} {Trajectory} {Generation} for {Bounding} {Flight}: {Combining} {Posture} {Control} and {Thrust} {Vectoring}},
	shorttitle = {Minimum {Time} {Trajectory} {Generation} for {Bounding} {Flight}},
	url = {https://ieeexplore.ieee.org/document/10178360},
	doi = {10.23919/ECC57647.2023.10178360},
	urldate = {2023-10-19},
	booktitle = {2023 {European} {Control} {Conference} ({ECC})},
	author = {Mandralis, Ioannis and Sihite, Eric and Ramezani, Alireza and Gharib, Morteza},
	month = jun,
	year = {2023},
	pages = {1--7},
}

@article{sihite_multi-modal_2023,
	title = {Multi-{Modal} {Mobility} {Morphobot} ({M4}) with appendage repurposing for locomotion plasticity enhancement},
	volume = {14},
	copyright = {2023 The Author(s)},
	issn = {2041-1723},
	url = {https://www.nature.com/articles/s41467-023-39018-y},
	doi = {10.1038/s41467-023-39018-y},
	language = {en},
	number = {1},
	urldate = {2023-07-04},
	journal = {Nature Communications},
	author = {Sihite, Eric and Kalantari, Arash and Nemovi, Reza and Ramezani, Alireza and Gharib, Morteza},
	month = jun,
	year = {2023},
	note = {Number: 1
Publisher: Nature Publishing Group},
	pages = {3323},
}

@inproceedings{liang_rough-terrain_2021,
	title = {Rough-{Terrain} {Locomotion} and {Unilateral} {Contact} {Force} {Regulations} {With} a {Multi}-{Modal} {Legged} {Robot}},
	doi = {10.23919/ACC50511.2021.9483189},
	booktitle = {2021 {American} {Control} {Conference} ({ACC})},
	author = {Liang, Kaier and Sihite, Eric and Dangol, Pravin and Lessieur, Andrew and Ramezani, Alireza},
	month = may,
	year = {2021},
	note = {ISSN: 2378-5861},
	pages = {1762--1769},
}

@inproceedings{sihite_optimization-free_2021,
	title = {Optimization-free {Ground} {Contact} {Force} {Constraint} {Satisfaction} in {Quadrupedal} {Locomotion}},
	doi = {10.1109/CDC45484.2021.9683155},
	booktitle = {2021 60th {IEEE} {Conference} on {Decision} and {Control} ({CDC})},
	author = {Sihite, Eric and Dangol, Pravin and Ramezani, Alireza},
	month = dec,
	year = {2021},
	note = {ISSN: 2576-2370},
	pages = {713--719},
}

@inproceedings{sihite_efficient_2022,
	title = {Efficient {Path} {Planning} and {Tracking} for {Multi}-{Modal} {Legged}-{Aerial} {Locomotion} {Using} {Integrated} {Probabilistic} {Road} {Maps} ({PRM}) and {Reference} {Governors} ({RG})},
	doi = {10.1109/CDC51059.2022.9992754},
	booktitle = {2022 {IEEE} 61st {Conference} on {Decision} and {Control} ({CDC})},
	author = {Sihite, Eric and Mottis, Benjamin and Ghanem, Paul and Ramezani, Alireza and Gharib, Morteza},
	month = dec,
	year = {2022},
	note = {ISSN: 2576-2370},
	pages = {764--770},
}

@article{grizzle_progress_nodate,
	title = {Progress on {Controlling} {MARLO}, an {ATRIAS}-series {3D} {Underactuated} {Bipedal} {Robot}},
	language = {en},
	author = {Grizzle, J W and Ramezani, A and Buss, B and Griﬃn, B and Hamed, K Akbari and Galloway, K S},
}

@article{dangol_control_2021,
	title = {Control of {Thruster}-{Assisted}, {Bipedal} {Legged} {Locomotion} of the {Harpy} {Robot}},
	volume = {8},
	issn = {2296-9144},
	url = {https://www.frontiersin.org/articles/10.3389/frobt.2021.770514},
	urldate = {2023-05-17},
	journal = {Frontiers in Robotics and AI},
	author = {Dangol, Pravin and Sihite, Eric and Ramezani, Alireza},
	year = {2021},
}

@inproceedings{dangol_feedback_2020-1,
	title = {Feedback design for {Harpy}: a test bed to inspect thruster-assisted legged locomotion},
	volume = {11425},
	shorttitle = {Feedback design for {Harpy}},
	url = {https://www.spiedigitallibrary.org/conference-proceedings-of-spie/11425/1142507/Feedback-design-for-Harpy--a-test-bed-to-inspect/10.1117/12.2558284.full},
	doi = {10.1117/12.2558284},
	urldate = {2023-05-17},
	booktitle = {Unmanned {Systems} {Technology} {XXII}},
	publisher = {SPIE},
	author = {Dangol, Pravin and Ramezani, Alireza},
	month = may,
	year = {2020},
	pages = {49--55},
}

@article{kim_bipedal_2021,
	title = {A bipedal walking robot that can fly, slackline, and skateboard},
	volume = {6},
	url = {https://www.science.org/doi/full/10.1126/scirobotics.abf8136},
	doi = {10.1126/scirobotics.abf8136},
	number = {59},
	urldate = {2022-10-30},
	journal = {Science Robotics},
	author = {Kim, Kyunam and Spieler, Patrick and Lupu, Elena-Sorina and Ramezani, Alireza and Chung, Soon-Jo},
	month = oct,
	year = {2021},
	note = {Publisher: American Association for the Advancement of Science},
	pages = {eabf8136},
}

@article{tobalske_aerodynamics_2007,
	title = {Aerodynamics of wing-assisted incline running in birds},
	volume = {210},
	issn = {0022-0949},
	url = {https://doi.org/10.1242/jeb.001701},
	doi = {10.1242/jeb.001701},
	number = {10},
	urldate = {2022-10-03},
	journal = {Journal of Experimental Biology},
	author = {Tobalske, Bret W. and Dial, Kenneth P.},
	month = may,
	year = {2007},
	pages = {1742--1751},
}

@article{dial_wing-assisted_2003-1,
	title = {Wing-{Assisted} {Incline} {Running} and the {Evolution} of {Flight}},
	volume = {299},
	url = {https://www.science.org/doi/full/10.1126/science.1078237},
	doi = {10.1126/science.1078237},
	number = {5605},
	urldate = {2022-10-03},
	journal = {Science},
	author = {Dial, Kenneth P.},
	month = jan,
	year = {2003},
	note = {Publisher: American Association for the Advancement of Science},
	pages = {402--404},
}

@inproceedings{apgar_fast_2018,
	title = {Fast {Online} {Trajectory} {Optimization} for the {Bipedal} {Robot} {Cassie}},
	isbn = {978-0-9923747-4-7},
	doi = {10.15607/RSS.2018.XIV.054},
	language = {en},
	urldate = {2019-12-12},
	booktitle = {Robotics: {Science} and {Systems} {XIV}},
	author = {Apgar, Taylor and Clary, Patrick and Green, Kevin and Fern, Alan and Hurst, Jonathan},
	month = jun,
	year = {2018},
}

@article{ramezani_performance_2014,
	title = {Performance {Analysis} and {Feedback} {Control} of {ATRIAS}, {A} {Three}-{Dimensional} {Bipedal} {Robot}},
	volume = {136},
	issn = {0022-0434},
	doi = {10.1115/1.4025693},
	language = {en},
	number = {2},
	urldate = {2020-05-09},
	journal = {Journal of Dynamic Systems, Measurement, and Control},
	author = {Ramezani, Alireza and Hurst, Jonathan W. and Akbari Hamed, Kaveh and Grizzle, J. W.},
	month = mar,
	year = {2014},
	note = {Publisher: American Society of Mechanical Engineers Digital Collection},
}

@inproceedings{buss_preliminary_2014,
	address = {Chicago, IL, USA},
	title = {Preliminary walking experiments with underactuated {3D} bipedal robot {MARLO}},
	isbn = {978-1-4799-6934-0 978-1-4799-6931-9},
	doi = {10.1109/IROS.2014.6942907},
	language = {en},
	urldate = {2020-01-04},
	booktitle = {2014 {IEEE}/{RSJ} {International} {Conference} on {Intelligent} {Robots} and {Systems}},
	publisher = {IEEE},
	author = {Buss, Brian G. and Ramezani, Alireza and Akbari Hamed, Kaveh and Griffin, Brent A. and Galloway, Kevin S. and Grizzle, Jessy W.},
	month = sep,
	year = {2014},
	pages = {2529--2536},
}

@article{koolen_design_2016,
	title = {Design of a {Momentum}-{Based} {Control} {Framework} and {Application} to the {Humanoid} {Robot} {Atlas}},
	volume = {13},
	issn = {0219-8436},
	url = {https://www.worldscientific.com/doi/10.1142/S0219843616500079},
	doi = {10.1142/S0219843616500079},
	number = {01},
	urldate = {2020-05-29},
	journal = {International Journal of Humanoid Robotics},
	author = {Koolen, Twan and Bertrand, Sylvain and Thomas, Gray and de Boer, Tomas and Wu, Tingfan and Smith, Jesper and Englsberger, Johannes and Pratt, Jerry},
	month = mar,
	year = {2016},
	note = {Publisher: World Scientific Publishing Co.},
	pages = {1650007},
}

@article{murphy_littledog_2011,
	title = {The {LittleDog} robot},
	volume = {30},
	issn = {0278-3649},
	doi = {10.1177/0278364910387457},
	language = {en},
	number = {2},
	urldate = {2020-04-15},
	journal = {The International Journal of Robotics Research},
	author = {Murphy, Michael P. and Saunders, Aaron and Moreira, Cassie and Rizzi, Alfred A. and Raibert, Marc},
	month = feb,
	year = {2011},
	note = {Publisher: SAGE Publications Ltd STM},
	pages = {145--149},
}

@inproceedings{wang_drc-hubo_2014,
	title = {{DRC}-hubo walking on rough terrains},
	doi = {10.1109/TePRA.2014.6869151},
	booktitle = {2014 {IEEE} {International} {Conference} on {Technologies} for {Practical} {Robot} {Applications} ({TePRA})},
	author = {Wang, Hongfei and Zheng, Yuan F. and Jun, Youngbum and Oh, Paul},
	month = apr,
	year = {2014},
	note = {ISSN: 2325-0534},
	pages = {1--6},
}

@misc{noauthor_robots_nodate,
	title = {Robots {\textbar} {Boston} {Dynamics}},
	url = {https://www.bostondynamics.com/robots},
	urldate = {2020-05-27},
}

@article{park_finite-state_2013,
	title = {A {Finite}-{State} {Machine} for {Accommodating} {Unexpected} {Large} {Ground}-{Height} {Variations} in {Bipedal} {Robot} {Walking}},
	volume = {29},
	issn = {1552-3098, 1941-0468},
	doi = {10.1109/TRO.2012.2230992},
	language = {en},
	number = {2},
	urldate = {2020-01-04},
	journal = {IEEE Transactions on Robotics},
	author = {Park, Hae-Won and Ramezani, Alireza and Grizzle, J. W.},
	month = apr,
	year = {2013},
	pages = {331--345},
}

\end{document}